\documentclass[runningheads]{llncs}
\usepackage{graphicx}
\usepackage{amsmath,amssymb} % define this before the line numbering.
\usepackage{color}
\usepackage{multirow}
\usepackage[width=122mm,left=12mm,paperwidth=146mm,height=193mm,top=12mm,paperheight=217mm]{geometry}

\begin{document}
% \linenumbers
\pagestyle{headings}
\mainmatter
\def\ECCV18SubNumber{***}  % Insert your submission number here

\title{MACNet: Multi-scale Atrous Convolution Networks for Food Places Classification in Egocentric Photo-streams} % Replace with your title

\titlerunning{EPIC@ECCV-18 submission ID 10}

\authorrunning{EPIC@ECCV-18 submission ID 10}

\author{Md. Mostafa Kamal Sarker\inst{1,\thanks{Corresponding Author: mdmostafakamal.sarker@urv.cat}} \and Hatem A. Rashwan\inst{1} \and Estefania Talavera \inst{3} \and Syeda Furruka Banu\inst{2} \and Petia Radeva\inst{3} \and Domenec Puig\inst{1}}

\institute{DEIM, Rovira i Virgili University, 43007 Tarragona, Spain.
\and ETSEQ, Rovira i Virgili University, 43007 Tarragona, Spain.
\and Department of Mathematics, University of Barcelona, 08007 Barcelona, Spain.
}

\maketitle

\begin{abstract}
First-person (wearable) camera continually captures unscripted interactions of the camera user with objects, people, and scenes reflecting his personal and relational tendencies. One of the preferences of people is their interaction with food events. The regulation of food intake and its duration has a great importance to protect against diseases. Consequently, this work aims to develop a smart model that is able to determine the recurrences of a person on food places during a day. This model is based on a deep end-to-end model for automatic food places recognition by analyzing egocentric photo-streams. In this paper, we apply multi-scale Atrous convolution networks to extract the key features related to food places of the input images. The proposed model is evaluated on an in-house private dataset called ``EgoFoodPlaces''. Experimental results shows promising results of food places classification recognition in egocentric photo-streams.
\end{abstract}

\section{Introduction}
The interest at lifelogging devices, such as first-person (wearable) cameras, being able to collect daily user information is recently increased. These cameras capable of frequently capturing images that record visual information of our daily life known as ``visual lifelogging'' in order to create a visual diary with activities of first-person life with unprecedented details~\cite{bolanos2017toward}. Since, the wearable camera can collect a huge number of images by non-stop image collection capacity (1-4 per minute, 1K-3K per day and 500K-1000K per year). The analysis of these egocentric photo-streams (images) can improve the people lifestyle; by analyzing social pattern characterization ~\cite{aghaei2018towards} and social interactions~\cite{aghaei2015towards}, as well as generating storytelling of first-person days~\cite{bolanos2017toward}. In addition, the analysis of these images can greatly affect on human behaviors, habits, and even health~\cite{grimm2011genetics}. One of the personal tendencies of people is food events that can badly affected on their health. For instance, some people can eat more if they see and senses (e.g. smell) food that constantly feel them hungry immediately~\cite{kemps2014exposure,de2012food}. Thus, monitoring and determining the duration of food intakes will help to improve people food behaviour.   

\begin{figure}[!t]
\centering
\includegraphics[width=0.9\textwidth]{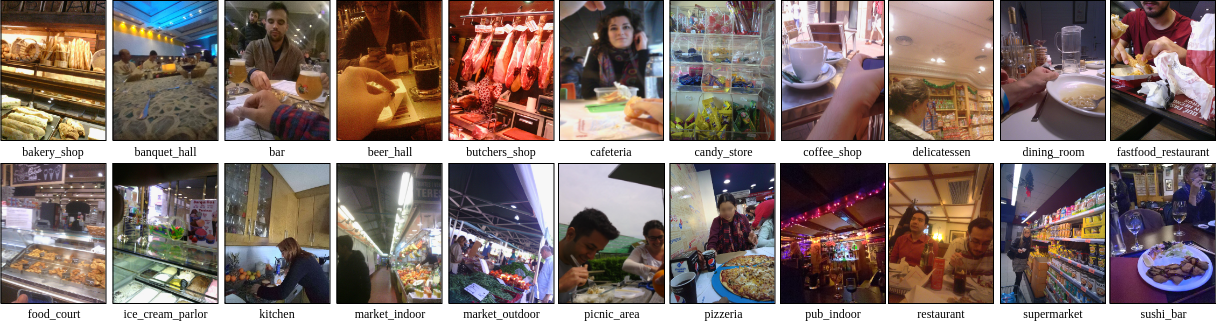}
\caption{Examples of images of food places from an in-house private EgoFoodPlaces dataset. EgoFoodPlaces is captured by 12 different users in different food places using the Narrative Clip camera. EgoFoodPlaces is employed to evaluate the proposed MACNet model for food places recognition.}
\label{fig:figure1}
% \vspace*{-2mm}
\end{figure}

\begin{figure}[!h]
\centering
\includegraphics[width=0.9\textwidth]{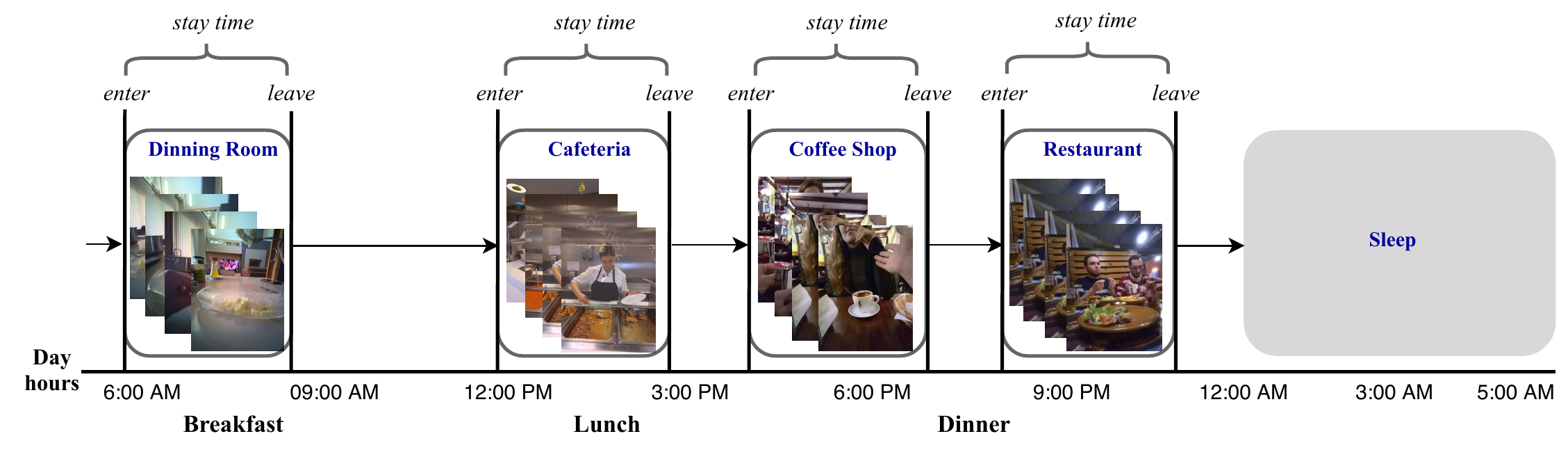}
\caption{Examples of commonly spending time in food places everyday.}
\label{fig:figure2}
\end{figure}

The motivation behind this research is twofold. Firstly, using a wearable camera is to capture images related to food places, where the users are engaged within foods (see Fig.~\ref{fig:figure1}). Consequently, these images of visual lifelogging can give a unique opportunity to work on food pattern analysis from the first-person viewpoint. Secondly, the analysis of everyday information (entering, leaving and stay time, see in Fig.~\ref{fig:figure2}) of visited food places can enable a novel health care application that can help to analyze the food eating patterns of people and prevent the diseases related to food, like obesity, diabetes and heart diseases.

% \vspace*{-5mm}

 Early work of places or scene recognition in conventional images was mainly motivated by two large scale places or scene datasets, i.e., Places2~\cite{zhou2014learning} and SUN397~\cite{xiao2010sun}) with millions of labeled images. The semantic classes of these datasets are defined by their labels by representing the entry-level of an environment. The images of datasets were collected form the internet with a large diversity. However, the two datasets failed to record the real involvement of first-person with food environment and the characterization of the first-person activity. In turn, wearable cameras can able to capture the scenes from a more intimate perspective by its ego-vision system. Thus, we built a new in-house private dataset, so-called ``EgoFoodPlaces'', with details involvement information of places that can help to classify the food places or environment to solve the first-person food pattern characterization.
 
 With diversity of food places (cafeterias, bars, restaurants, ..etc.) traditional methods of feature extraction (e.g., HOG and SIFT) and classification (e.g., Support Vector Machine (SVM) and Neural Network (NN))~\cite{moran2006looking} are not sufficient to deal with this complex problem of food places recognition. Thus, this paper aims to use deep learning models (e.g., Convolutional Neural Network (CNN)) that will help us to automatically select and extract key features and also to construct new ones for different food places. One of recent architectures of deep networks used for classification and segmentation tasks is Atrous Convolution Networks proposed in~\cite{atrous2018}. That networks can encode contextual information by using filters or pooling operations at multiple rates with different sizes of neighbourhoods. Thus, in this paper, we propose to use these networks in our deep model to improve the classification rate with ResNet networks. In addition to detect important structures as well as small details of the input images, we rescale the input images in a multi-scale space (i.e., a pyramid of images with different resolutions).

The main contributions of this work is summarized as follows: 
\begin{itemize}
\vspace*{-2mm}
  \item [$\bullet$] Introduce a new dataset developed by lifelogging camera for food places classification, named EgoFoodPlaces.
  \item [$\bullet$] Proposed a new deep network architecture based on multi-scale Atrous convolutional networks~\cite{atrous2018} for improving classification rate of food places in egocentric photo-streams. 
\vspace*{-2mm}
 \end{itemize}

The paper is organized as follows. Section 2 explores the proposed approach. In turn, Section 3 describes our in-house  dataset and demonstrate the experimental results and  discussions. Finally, conclusion and future work are shown in section 4.

\section{Proposed Approach}
\begin{figure}[htp]
    \vspace*{-2mm}
	\centering
	\includegraphics[width=0.9\textwidth]{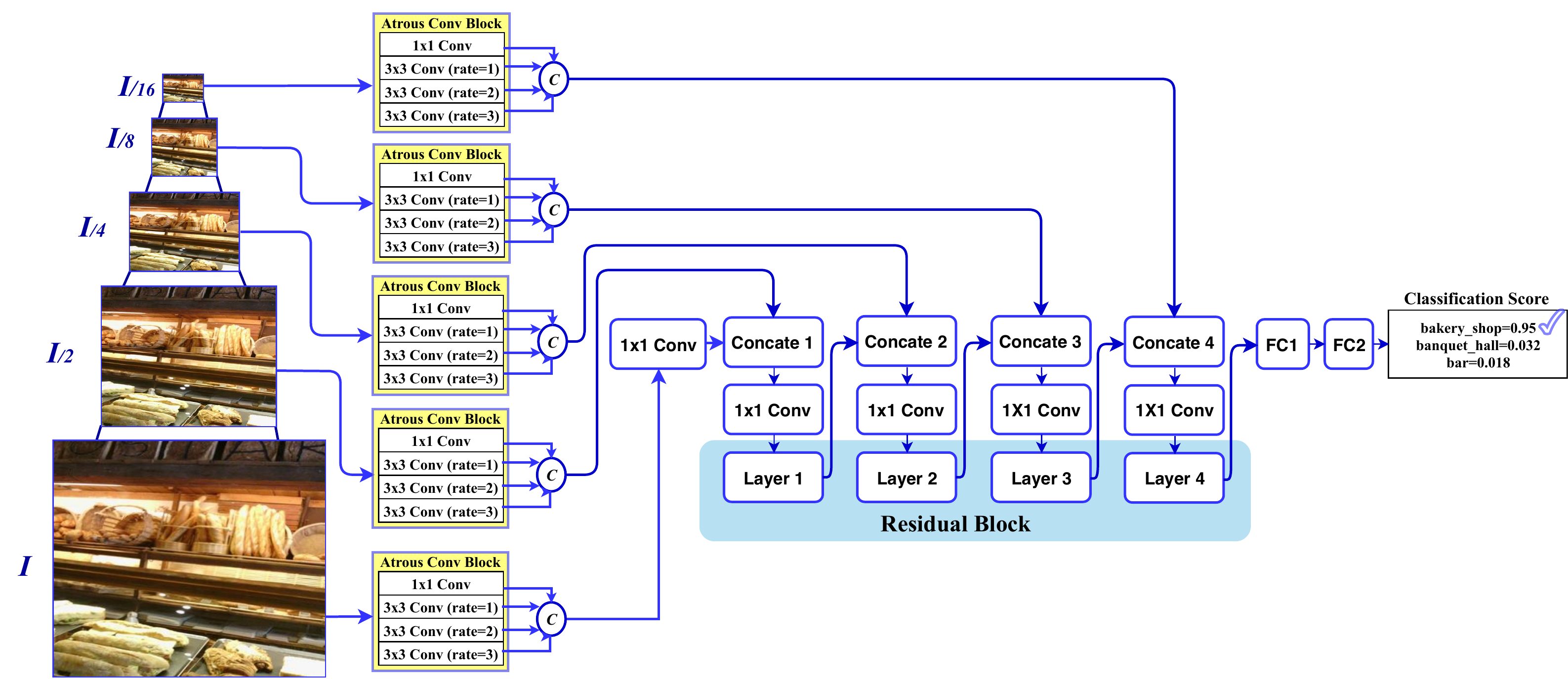}
	\caption{Architecture of our proposed model for food places classification. }
	\label{fig:figure3}
    \vspace*{-2mm}
\end{figure}

The proposed deep model, MACNet, is based on multi-scale Atrous convolutional networks for extracting the key patterns of food places in the input egocentric photo-streams. The multi-scale features are used to fine tuning four layers of a pre-trained ResNet-101 model as shown in figure~\ref{fig:figure3}.

%\subsection{Network Architecture}
The input images are scaled to five resolutions (i.e., the original size and four different resolutions) as shown in figure~\ref{fig:figure3}. The five images with different resolutions feeds to Atrous convolutional networks~\cite{atrous2018}. In MACNet, five blocks of Atrous convolutional network with three different rates per block are used to extract the key features of an input image. Atrous convolutional network allows us to explicitly extract features with different scales. In addition, it adjusts filter’s size with the rate value in order to capture multi-scale information, generalizes standard convolution operation. We used $3\times 3$ kernels in all blocks with different rate values set to  1, 2 and 3. More details about these networks presented in~\cite{chen2017rethinking} and \cite{atrous2018}.

Following, four pre-trained ResNet-101 blocks are then used to extract $256$, $512$, $1024$ and $2048$ feature maps, respectively as shown in figure~\ref{fig:figure3}. The four ResNet-101 layers are with stride $2.0$. Thus, the final output size of the last ResNet block is $1/16$ of the input image size. Indeed, each ResNet is corresponding to a resolution level in the image pyramid. Each output of the five Atrous network blocks is followed by a pointwise convolution (i.e., $1\times 1$ convolution) to reduce the computation complexity and the number of channels to be compatible with the input channels accepted by the corresponding ResNet layer. All Atrous convolutional networks and $1\times 1$ convolution are randomly initialized.

The output of the fourth ResNet layer feeds to a fully connected layer with $1024$ neurons followed by another fully connected layer with $512$ neurons. A dropout function with $0.5$ is used for reducing overfitting in the two fully connected layers. A ReLU function is also used as an activation function for the first fully connected layer. In turn, a softmax function (i.e., normalized exponential function) is finally utilized as a logistic function for producing the final probability of the input image to each class. The two fully connected layers are randomly initialized.

%\subsection{Loss Function}
In this paper, we constructed new dataset from egocentric images. In our dataset ``EgoFoodPlaces'', the numbers of instances with different labels are very unbalanced. To deal with this issue, we used a weighted categorical cross-entropy function. For the cross-entropy with a multi-class classification, we calculate a separate loss for each class label and then then sum the result as:
\begin{equation}
\ell_i = - \sum_{j=1}^N w_{y_j}y_j \log(\hat{y_j}),
\end{equation}
where $N$ is the number of instances, $y_j$ is the actual label of the $j_th$ instance, $\hat{y_j}$ is the prediction score, and $w_{y_j}$, the loss weight of the label $y_j$, is defined as:
\begin{equation}
w_{y_j}= 1 - \frac{N_{y_j}}{N},
\label{weight}
\end{equation}
where $N_{y_j}$ refers to the number of instances per label $y_j$.

\section{Experimental Results}

\subsection{Datasets}
In this work we introduce to ``EgoFoodPlaces'', a new egocentric photo-streams dataset that devolved by $12$ users by using wearable camera (narrative clip 2~\footnote{http://getnarrative.com/}, which has an image resolution of 720p and 1080p by a 8-megapixel camera with an 86-degree field of view and capable of record about $4,000$ photos or $80$ minutes of 1080p video at 30fps). Figure~\ref{fig:figure1} shows some example images from the EgoFoodPlaces dataset. It is composed by egocentric photo-streams describing the users daily food related activities (preparing, eating, buying, etc). Some images are used in our data, EgoFoodPlaces, from the EDUB-Seg dataset~\cite{dimiccoli2017sr}. 

The first-person used the camera fixed to his chest from morning to night before sleeping. Figure ~\ref{fig:figure2} shows the day hours for capturing the images. Every frames of a photo-stream is recording first-person activities, which is very helpful to analyze different pattern of first-person lifestyle. However, the captured images have different challenges, such as background variation, lighting change, and handling objects sometimes occluded during the photo-stream. In addition, the constructed dataset has unbalanced classes. However, it is not possible to make it as a balanced dataset by reducing images from other classes, since some classes have very small number of images. The classes with few images are related to some food places that do not have rich visual information (e.g., candy store) or the users do not spend much time at there (e.g., butchers shop). In turn, we have very large number of images are of visited places with rich visual information that refer to daily contexts (e.g., kitchen, supermarket), or of places, where we send more times (e.g., restaurant). We labelled our dataset manually by taking the reference labels related to food places from the Places2 dataset~\cite{zhou2014learning}. Since, some of the classes related to food places from the Places2 dataset~\cite{zhou2014learning} are not available (e.g., beer garden). Therefore, we excluded these classes from EgoFoodPlaces. 

Twenty-two classes of food places are described in our dataset as shown in Table~\ref{tab:dataset}. We have split EgoFoodPlaces into three sets: train, validation and test. The images of each set were not randomly choose to avoid of taking similar images from the same events. Thus, we split the dataset based on food event information. The events represents the entry and exit image frame from the places visited. This can make the dataset more robust to train and validate our model.

\begin{table}[h!]
\centering
\caption{The distribution of images per class in the EgoFoodPlaces dataset.}
\resizebox{\textwidth}{!}{%
\begin{tabular}{|c|c|c|c|c|c|c|c|c|c|c|c|c|c|}
\hline
 & \multicolumn{2}{c|}{Train} & \multicolumn{2}{c|}{Val} & \multicolumn{2}{c|}{Test} &  & \multicolumn{2}{c|}{Train} & \multicolumn{2}{c|}{Val} & \multicolumn{2}{c|}{Test} \\ \hline
Classes & images & events & images & events & images & events & Classes & images & events & images & events & images & events \\ \hline
bakery\_shop & 96 & 15 & 15 & 3 & 28 & 4 & food\_court & 161 & 6 & 37 & 2 & 06 & 1 \\ \hline
banquet\_hall & 203 & 1 & 52 & 1 & 96 & 1 & ice\_cream\_parlor & 70 & 4 & 12 & 2 & 25 & 1 \\ \hline
bar & 1121 & 23 & 137 & 5 & 374 & 6 & kitchen & 2701 & 81 & 389 & 13 & 743 & 23 \\ \hline
beer\_hall & 296 & 1 & 62 & 1 & 318 & 1 & market\_indoor & 644 & 15 & 97 & 3 & 163 & 4 \\ \hline
butchers\_shop & 251 & 4 & 11 & 1 & 15 & 1 & market\_outdoor & 1271 & 11 & 13 & 2 & 104 & 3 \\ \hline
cafeteria & 1238 & 23 & 141 & 5 & 310 & 6 & picnic\_area & 659 & 4 & 89 & 2 & 173 & 1 \\ \hline
candy\_store & 172 & 4 & 26 & 2 & 55 & 1 & pizzeria & 1022 & 3 & 125 & 1 & 265 & 1 \\ \hline
coffee\_shop & 1662 & 29 & 210 & 5 & 441 & 8 & pub\_indoor & 342 & 7 & 60 & 1 & 109 & 2 \\ \hline
delicatessen & 652 & 6 & 29 & 2 & 05 & 1 & restaurant & 4198 & 29 & 481 & 5 & 1044 & 8 \\ \hline
dining\_room & 2481 & 73 & 326 & 12 & 832 & 21 & supermarket & 3019 & 70 & 477 & 10 & 827 & 20 \\ \hline
fastfood\_restaurant & 858 & 14 & 102 & 2 & 217 & 4 & sushi\_bar & 1151 & 7 & 195 & 1 & 296 & 2 \\ \hline
\end{tabular}}
\label{tab:dataset}
\end{table}

\subsection{Experimental Setup}
The proposed model is implemented on PyTorch\cite{paszke2017pytorch}: an open source deep learning library. For the optimization method, we used the Stochastic Gradient Descent (SGD)~\cite{gulcehre2017robust} with momentum of 0.9 and weight decay of 0.0005. For adjusting learning rate depending on first and second order moments of the gradient, we used a ``step'' learning rate policy~\cite{sebag2017stochastic} and we selected a base learning rate of 0.001 and the step is 20.  In order to increase the number of images related to a class having few images, we used data augmentation. For data augmentation, we performed random crop, image brightness and contrast change with 0.2 and 0.1, respectively. We also use random affine transform between the angle of -20 and 20, image translation of 0.5, random scale between 0.5 and 1.0, and random rotation of 10 degrees. The optimized batchsize is set to $32$ for training and the number of epochs is set to $100$. All the experiments are executed on NVIDIA TITAN X with 12GB memory taking around 20 hours to train the network. All these parameters are used for all tested methods in our experiments.

\subsection{Evaluation metrics}
Since the constructed dataset, EgoFoodPlaces, is highly imbalanced, the classification performance of all tested methods was assessed by not only using the accuracy, but also using other three evolution measures: precision, recall, and F1-score. 

\subsection{Comparison with classification methods}
Three different CNN architectures, specifically the VGG16, InceptionV3, and ResNet 50, are used in a comparison to assess our proposed model, MACNet.

\textbf{VGG-16:} We fine-tuned a VGG-16 network proposed in~\cite{schussler2017} in the all 16 layers were back-propagated, and the SGD optimization method used.

\textbf{ResNet50:} The ResNet50 network proposed in~\cite{he2016deep} was fine-tuned and was optimized using SGD.

\textbf{InceptionV3:} The InceptionV3 network proposed in~\cite{chen2017rethinking} was also fine-tuned with SGD
as an optimization method.

\subsection{Results and Discussions}
We compared the performances of VGG16, ResNet50 and InceptionV3 to our proposed model, MACNet as shown in Table ~\ref{table2}. MCANet yielded an average of Precision of $72\%$, Recall of $60\%$ and F1-score of $65\%$ with the validation set, and about $70\%$, $57\%$ and $63\%$, respectively with the test set. Our experiments demonstrated that the food places classification scores obtained with MACNet are better than the scores of the three test models on both validation and test set. However, InceptionV3 provided acceptable results with around $61\%$, $50\%$ and $55$ with both validation and test sets. In turn, VGG16 yielded the worst scores among the four tested method. This means that the MACNet based on multi-scale Atrous convolutional networks can be able to improve the classification of food places in egocentric photo-images.

\begin{table}[h!]
\centering
\caption{The average Precision, Recall and F1-score of both validation and test sets of the EgoFoodPlaces dataset with VGG16, ResNet50, InceptionV3 and the proposed MACNet model.}
\label{table2}
\scalebox{0.8}{
% \resizebox{\textwidth}{!}{%
\begin{tabular}{|c|c|c|c|c|c|c|}
\hline
\multirow{2}{*} {Models}  & \multicolumn{3}{c|}{Validation} & \multicolumn{3}{c|}{Test}     \\ \cline{2-7}                    	         
             & Precision  & Recall    & $F_{1}$  score  & Precision  & Recall     & $F_{1}$  score \\ \hline \hline
VGG16        &    38.12   &  25.06     &  30.24         &   36.46    &  24.85    &   29.55           \\ \hline
ResNet50     &    61.30   &  49.04     &  54.48         &   59.07    &  47.44    &   52.62            \\ \hline
InceptionV3  &    63.91   &  52.13     &  57.42         &   61.39    &  50.51    &   55.42           \\ \hline
\textbf{MACNet}& \textbf{72.33}  & \textbf{59.53}  & \textbf{65.37}    & \textbf{69.54}   & \textbf{57.19}  &  \textbf{62.76}       \\ \hline
\end{tabular}%
}
\end{table}

Furthermore, the Top-1 and Top-5 accuracy of the three test models, VGG16, ResNet50 and InceptionV3, and the proposed MACNet model are shown in Table~\ref{tableTop15}. As shown, for the validation set, MACNet yielded more than a $10\%$ improvement in Top-1 accuracy with respect to the VGG-16 model, and around a $4\%$ improvment with respect to both ResNet50 and InceptionV3 models. Regarding to the test set, MACNet lead to a $3\%$ improvement to the three tested model.

\begin{table}[h!]
\centering
\caption{The average Top-1 and Top-5 accuracy of both validation and test sets of the EgoFoodPlaces dataset with VGG16, ResNet50, InceptionV3 and the proposed MACNet model.}
\label{tableTop15}
\scalebox{0.8}{
% \resizebox{\textwidth}{!}{%
\begin{tabular}{|c|c|c|c|c|}
\hline
\multirow{2}{*} {Models}  & \multicolumn{2}{c|}{Validation} & \multicolumn{2}{c|}{Test}     \\ \cline{2-5}  
& Top-1  & Top-5   & Top-1  & Top-5 \\ \hline \hline
VGG16        &53.93   &83.98   &49.20   &81.07  \\ \hline
ResNet50     &61.31   &85.48    &55.38  &84.95  \\ \hline
InceptionV3  &60.82   &88.22    &54.76  &85.60  \\ \hline
\textbf{MACNet}& \textbf{64.80}  & \textbf{90.70}   & \textbf{58.47}  &  \textbf{86.78}       \\ \hline

\end{tabular}%
}
\end{table}

\vspace*{-4mm}
\begin{figure}[htp]
	\centering
	\includegraphics[width=0.95\textwidth]{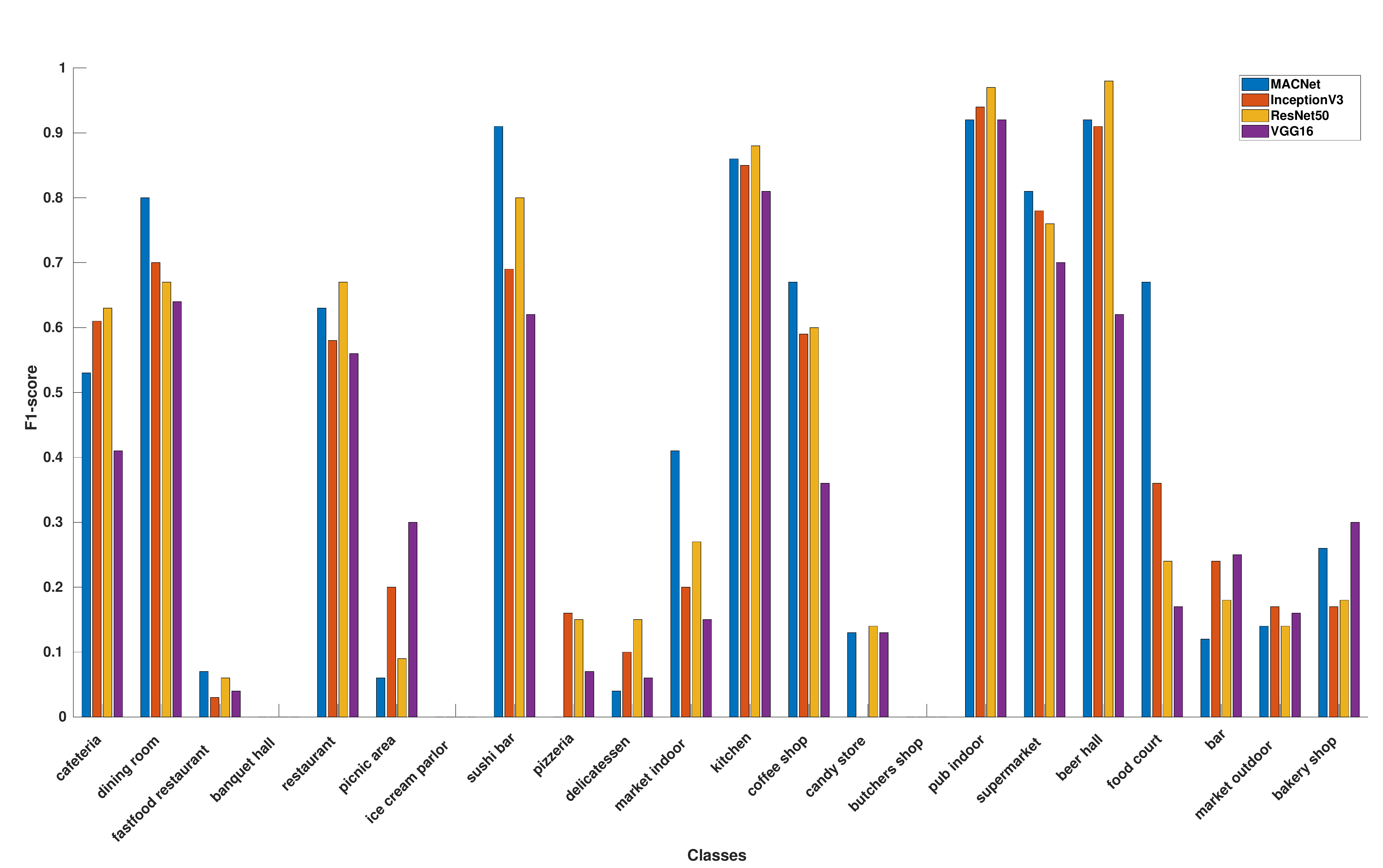}\\
	\includegraphics[width=0.95\textwidth]{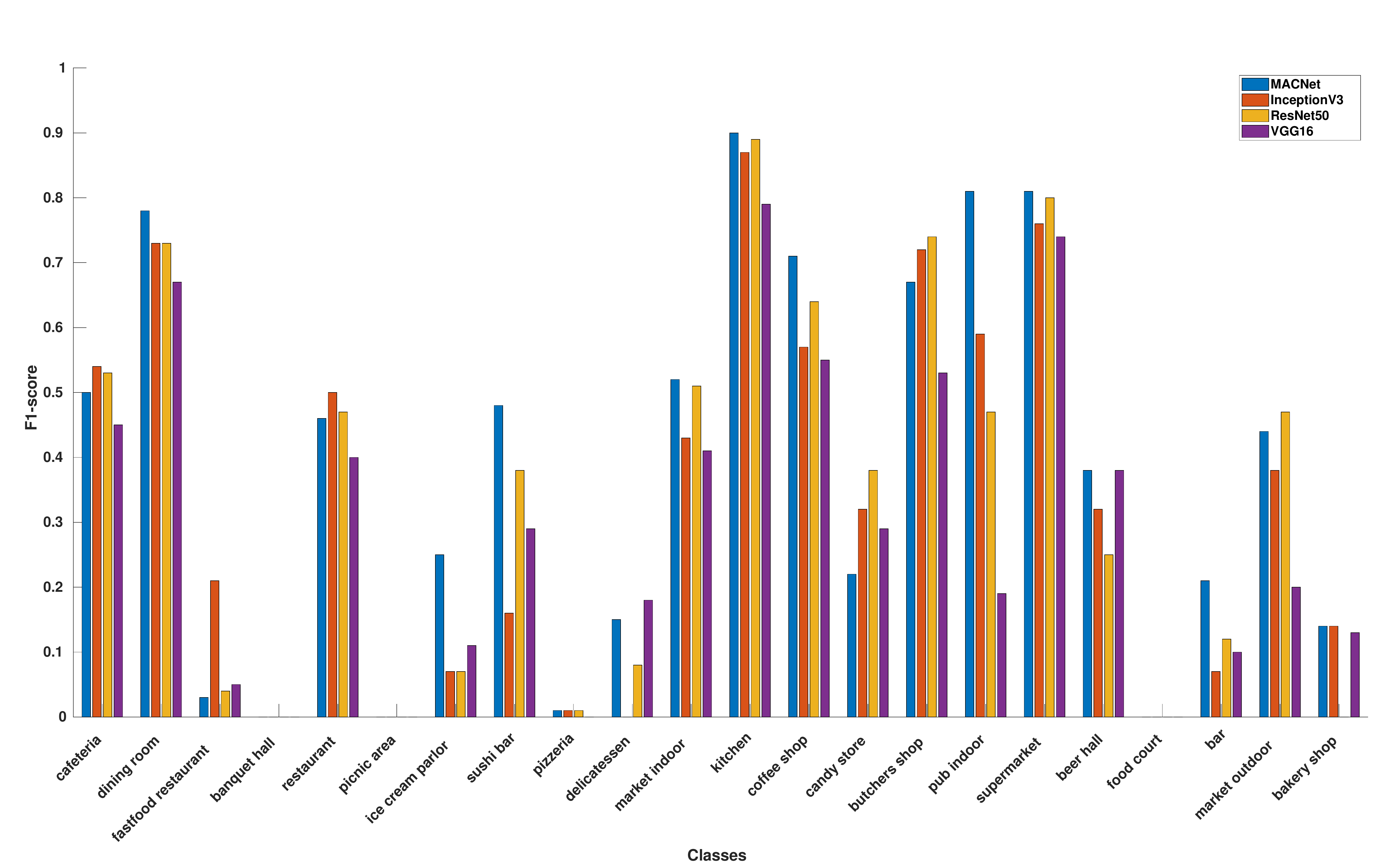}
	\caption{The resulted of F1-score of the (Top) validation set (down) test set of the EgoFoodPlaces dataset with three methods VGG16, ResNet50, InceptionV3 and the proposed MACNet model.}
	\label{fig:figure4}
 	\vspace*{-4mm}
\end{figure}

Figure~\ref{fig:figure4} shows the F1-score per class with the four tested methods over 22 classes of the validation and test set of EgoFoodPlaces. In the most classes (e.g., dining room, sushi bar, ice cream, coffee shop and food court), MACNet yielded a significant improvement of F1-score. In some cases (e.g., hall bar and pub indoor), ResNet50 provided better results than the other methods. In turn, VGG16 can classify the food places in the EgoFoodPlaces better then the other tested methods, such as picnic area and bakery shop. While, InceptionV3 did not outperform the other methods per class, however its average F1-score is better than VGG16 and ResNet50 and less than MACNet. Note that the zero values of F1-score shown in figure~\ref{fig:figure4} are related to the classes that have few images per class.

\begin{figure}[htp]
	\centering
	\includegraphics[width=0.9\textwidth]{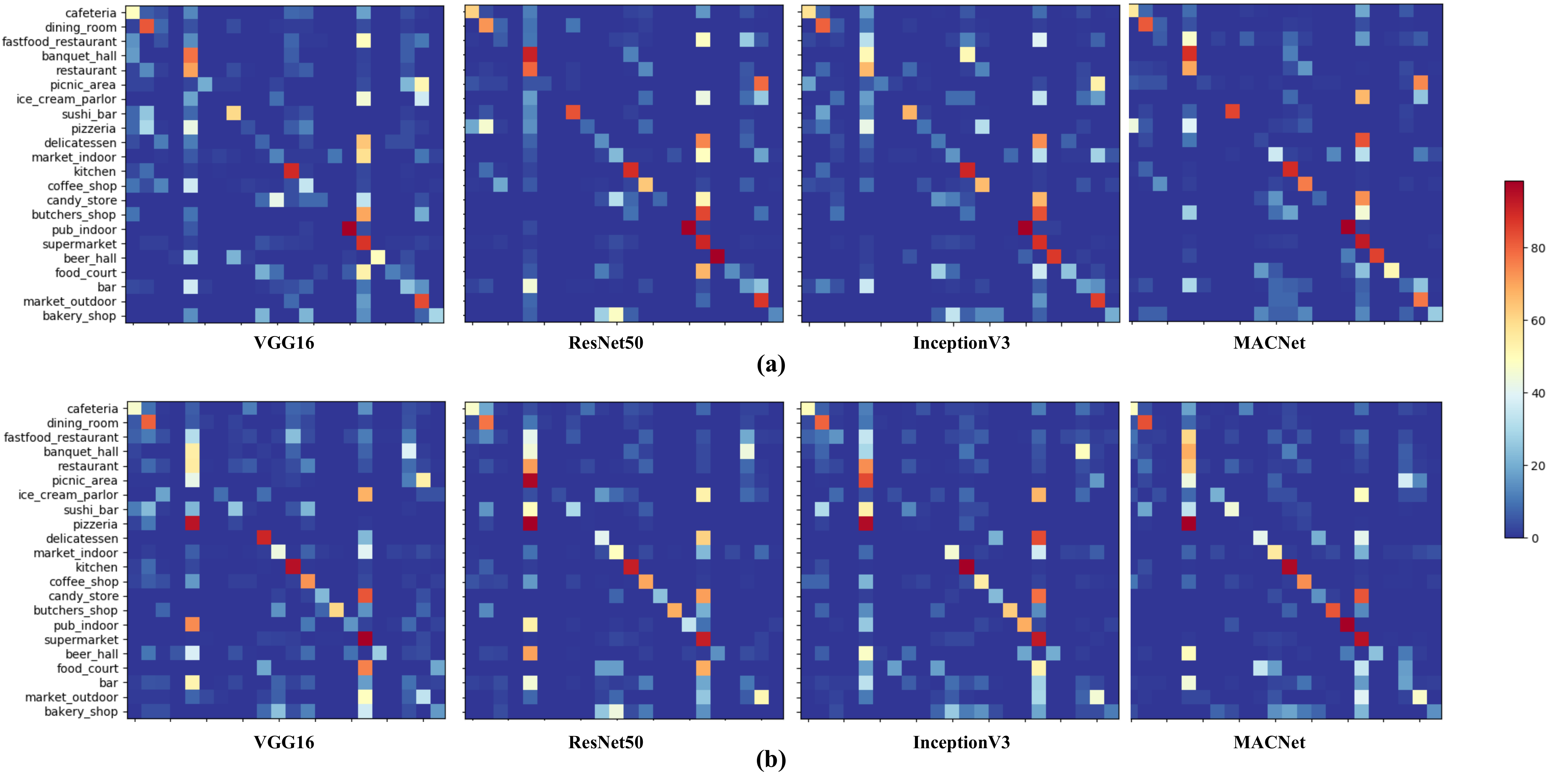}
	\caption{The confusion matrix of the (top) validation set and (down) test set of the EgoFoodPlaces dataset with three methods VGG16, ResNet50, InceptionV3 and the proposed MACNet model.}
	\label{fig:figure5}
 	\vspace*{-4mm}
\end{figure}

Moreover, the improvement over the overlapping classes can also be seen on the confusion matrices shown in figure~\ref{fig:figure5}. This means that the multi-scale Atrous convolutional networks improved the food places classification belonging to classes that score similar probabilities.

\section{Conclusions}

In this paper, we proposed a new architecture of deep model, MACNet for food places recognition from egocentric photo-streams. MACNet is based on multi-scale Atrous convoluitonal networks that fusing with four pre-tranied layers of ResNet-101 and two fully connected layers.
MACNet extracting features of different resolutions of an input image of first-person images. In addition, we constructed an in-house private egocentric photo-streams dataset containing 22 classes of food places, named EgoFoodPlaces. Experimental results on this dataset demonstrated
that the proposed approach achieve better performances
than a three common architecture of classification methods, VGG16, ResNet50 and InceptionV3. The proposed method achieved
an overall Top-5 accuracy around $86.78\%$ over the test set of EgoFoodPlaces. Future work aims to use the MACNet model with a complete framework for people food behaviour.

\bibliographystyle{splncs}
\bibliography{egbib}
\end{document}